# Ground Truth for training OCR engines on historical documents in German Fraktur and Early Modern Latin

Uwe Springmann
Ludwig-Maximilians-Universität München
uwe@springmann.net

Christian Reul
Universität Würzburg
christian.reul@uni-wuerzburg.de

Stefanie Dipper
Ruhr-Universität Bochum
dipper@linguistics.rub.de

Johannes Baiter
Bayerische Staatsbibliothek München
baiter@bsb-muenchen.de

In this paper we describe a dataset of German and Latin *ground truth* (GT) for historical OCR in the form of printed text line images paired with their transcription. This dataset, called *GT4HistOCR*, consists of 313,173 line pairs covering a wide period of printing dates from incunabula from the 15th century to 19th century books printed in Fraktur types and is openly available under a CC-BY 4.0 license. The special form of GT as line image/transcription pairs makes it directly usable to train state-of-the-art recognition models for OCR software employing recurring neural networks in LSTM architecture such as Tesseract 4 or OCRopus. We also provide some pretrained OCRopus models for subcorpora of our dataset yielding between 95% (early printings) and 98% (19th century Fraktur printings) character accuracy rates on unseen test cases, a Perl script to harmonize GT produced with different transcription guidelines, and give hints on how to construct GT for OCR purposes which has requirements that may differ from linguistically motivated transcriptions.

# 1 Introduction

The conversion of scanned images of printed historical documents into electronic text by means of OCR has recently made excellent progress, regularly yielding character recognition rates by individually trained models beyond 98% for even the earliest printed books (Springmann, Fink, and Schulz, 2015; Springmann and Fink, 2016; Springmann and Lüdeling, 2017; Springmann, Fink, and Schulz, 2016; Reul, Dittrich, and Gruner, 2017; Reul et al., 2017; Reul et al., 2018, see also this volume). This is due to (1) the application of recurrent neural networks with LSTM architecture to the field of OCR (Fischer et al., 2009; Breuel et al., 2013; Ul-Hasan and Breuel, 2013), (2) the availability of open source OCR engines which can be trained on specific scripts and fonts such as Tesseract[1] and OCRopus,[2] and (3) the possibility to train recognition models on real printed text lines as opposed to generating artifical line images from existing computer fonts (Breuel et al., 2013; Springmann et al., 2014).

What is missing, however, are robust pretrained recognition models applicable to a wide range of typographies spanning different fonts (such as Antiqua and Fraktur with long s), scripts and publication periods, which would yield a tolerable OCR result of >95% character recognition rate without the need of any specific training. Accurate ground truth and better individual OCR models could be constructed from the output of these pretrained models much more easily than by transcriptions from scratch. The feasibility to construct such mixed models able to generalize to previously unseen books that have not contributed to model training has been shown in Springmann and Lüdeling (2017) with diachronic German Fraktur printings (compare their Fig. 6 and Fig. 7): Character recognition rates of individual models quickly fall below 80% when applied to books printed with different fonts at different periods, whereas mixed models show an average rate of 95% (see their Table 2).

The construction of pretrained mixed models crucially depends on available ground truth data for a wide variety of historical documents. In this paper we describe training material of historical ground truth which has been collected and produced by us over the course of several years. The training of OCRopus models is described in detail in the CIS workshop on historical OCR[3] and the OCRoCIS tutorial.[4]

All of our ground truth is made available in the GT4HistOCR (*Ground Truth for Historical OCR*) dataset under a CC-BY 4.0 license in Zenodo (Springmann et al., 2018). The repository contains the compressed subcorpora, some pretrained mixed OCRopus models for subcorpora, and a Perl script which can be adapted to harmonize GT produced by different transcriptions guidelines in order to have a common pool of training data for mixed models.

In the following we describe our GT4HistOCR dataset and its constituent subcorpora (Sect. 2), mention other existing sources of historical GT which have not yet been mined for model construction (Sect. 3) together with a description of a crowdsourcing tool for GT production using public

[1] https://github.com/tesseract-ocr/tesseract/wiki/Training-Tesseract
[2] https://github.com/tmbdev/ocropy/wiki
[3] http://www.cis.uni-muenchen.de/ocrworkshop/program.html
[4] http://cistern.cis.lmu.de/ocrocis/tutorial.pdf

**Table 1:** Overview of the subcorpora of *GT4HistOCR*. For each subcorpus we indicate the number of books, the printing period, the number of lines, and the language.

| Sect. | Subcorpus | # Books | Period | # Lines | Language |
|---|---|---|---|---|---|
| 3.1 | Reference Corpus ENHG | 9 | 1476-1499 | 24,766 | ger |
| 3.2 | Kallimachos Corpus | 9 | 1487-1509 | 20,929 | ger, lat |
| 3.3 | Early Modern Latin | 12 | 1471-1686 | 10,288 | lat |
| 3.4 | RIDGES Fraktur | 20 | 1487-1870 | 13,248 | ger |
| 3.5 | DTA19 | 39 | 1797-1898 | 243,942 | ger |
| | | | **Sum:** | **313,173** | |

APIs of the Internet Archive (Sect. 3.1), make some remarks about transcription guidelines and their relevance to the production of GT for OCR purposes (Sect. 4), and end with a conclusion (Sect. 5).

## 2 The *GT4HistOCR* dataset

In the following we introduce the five subcorpora of our *GT4HistOCR* dataset (see Table 1). The transcription of these corpora was done manually (partly by students) and later checked and corrected by trained philologists within projects in which we participated:

The *Reference Corpus Early New High German*[5] is a DFG funded project, the *Kallimachos* corpus derives from work done in the BMBF funded Kallimachos project,[6] the *Early Modern Latin* corpus was produced during projects on OCR postcorrection funded by CLARIN and DFG,[7] *RIDGES*[8] has been built by students at HU Berlin as part of their studies in historical corpus linguistics and *DTA19* has been extracted from the DFG-funded *Deutsches Textarchiv* (DTA).[9] An overview of the contribution of these subcorpora to our dataset is shown in Table 1.

The text line images corresponding to the transcripted lines have been prepared and matched by us using OCRopus segmentations routines or, in the case of DTA19, the segmentation of ABBYY Finereader. The ground truth in the form of paired line images and their transcriptions are an excerpt from the books in a corpus.

Because the transcription guidelines for each subcorpus differ in the amount of typographical detail that has been recorded we chose not to construct corpora according to language or period by merging and harmonizing material from these subcorpora. However, because the directory containing the GT of each book is named with publishing year and book title, a user can remix our data and

[5] https://www.linguistics.ruhr-uni-bochum.de/ref/
[6] http://kallimachos.de
[7] http://www.cis.lmu.de/ocrworkshop/
[8] http://korpling.org/ridges/
[9] http://www.deutschestextarchiv.de/

```
File name: EarlyModernLatin/1564-Thucydides-Valla/00001.bin.png
```

fuerunt,tum uetuſtæ,tum ſequentiũ temporum.Non minimam

```
File name: EarlyModernLatin/1564-Thucydides-Valla/00001.gt.txt
```

fuerunt, tum uetuſtæ, tum ſequentiũ temporum. Non minimam construct new corpora according to his needs after the transcriptions have been harmonized. An example of a GT line pair is given in Fig. 1.

**Figure 1:** Example GT line pair of line image (upper line) and its transcription. A blank after each punctuation symbol has been added and the OCR model will consequently learn to map a punctuation symbol to the sequence punctuation, blank.

### 2.1 Incunabula from the *Reference Corpus Early New High German*

The Reference Corpus Early New High German (ENHG) is being created in an ongoing project which is part of a larger initiative with the goal of creating a diachronic reference corpus of German, starting with the earliest existing documents from Old High German and Old Saxon (750–1050), and including documents from Middle High German (1050–1350) and Middle Low German and Low Rhenish (1200-1650), up to Early New High German (1350–1650). The Reference Corpus Early New High German contains texts published between 1350 and 1650. From 1450 on, prints are included in the corpus besides manuscripts. The last part, 1550–1650, consists of prints only. The texts have been selected in a way as to represent a broad and balanced selection of available language data. The corpus contains texts from different time periods, language areas, and document genres (e.g. administrative texts, religious texts, chronicles). From the Reference Corpus Early New High German we got ground truth for the incunabula printings in Table 2. Specimen of line images which give an impression of the fonts are shown in Fig. 2. Full bibliographic details for these documents can be retrieved from the *Gesamtkatalog der Wiegendrucke*[10] via the GW number.

While in principle we would like to have as large a corpus as possible and reuse all transcriptions from 1450 up to 1650, the process of generating accurately segmented printed lines from scanned book pages and matching them to their corresponding transcriptions is still laborious. Because OCR ground truth for periods later than 1500 is provided in other subcorpora we just used the incunabula printings of the reference corpus.

We also wanted to explore the feasibility to construct a mixed model and test its predictive power for unseen works from this period. For the about 30,000 incunabula printings, about 2000 print shops (*officinae*) using about 6000 typesets have been identified in the *Typenrepertorium der Wiegendrucke*,[11] so a mixed model trained on only a few books might not generalize well to other incunabula printed in one of the many other and possibly much different fonts. On the other hand, even in this early period a divison of labour between punchcutters and printers took place and commercially successful printing types were available for sale (Carter, 1969), so it might be expected that not all 6,000 identified fonts employed in the print shops were totally different from each others.

[10] http://www.gesamtkatalogderwiegendrucke.de/GWEN.xhtml
[11] http://tw.staatsbibliothek-berlin.de/

**Table 2:** The Early New High German incunabulum corpus. Given are the printing year, the GW number, the short title, the number of ground truth lines for training and evaluation, and the character recognition rate (CRR) in % of a mixed model trained on all other books.

| Year | GW | (Short) Title | # Lines | CRR |
|---|---|---|---|---|
| 1476 | M51549 | Historij | 3160 | 96.11 |
| 1478 | 04307 | Biblia | 2745 | 91.90 |
| 1485 | M09766 | Gart der Gesuntheit | 2520 | 96.37 |
| 1486 | M45593 | Eunuchus | 3403 | 92.61 |
| 1486 | 5077 | Jherusalem | 2232 | 97.83 |
| 1490 | 10289 | Pfarrer vom Kalenberg | 2503 | 98.07 |
| 1490 | 5793 | Leben und Sitten | 3099 | 93.59 |
| 1497 | 5593 | Cirurgia | 3476 | 96.16 |
| 1499 | 6688 | Cronica Coellen | 1628 | 95.98 |
| | | **Sum:** | **24,766** | |

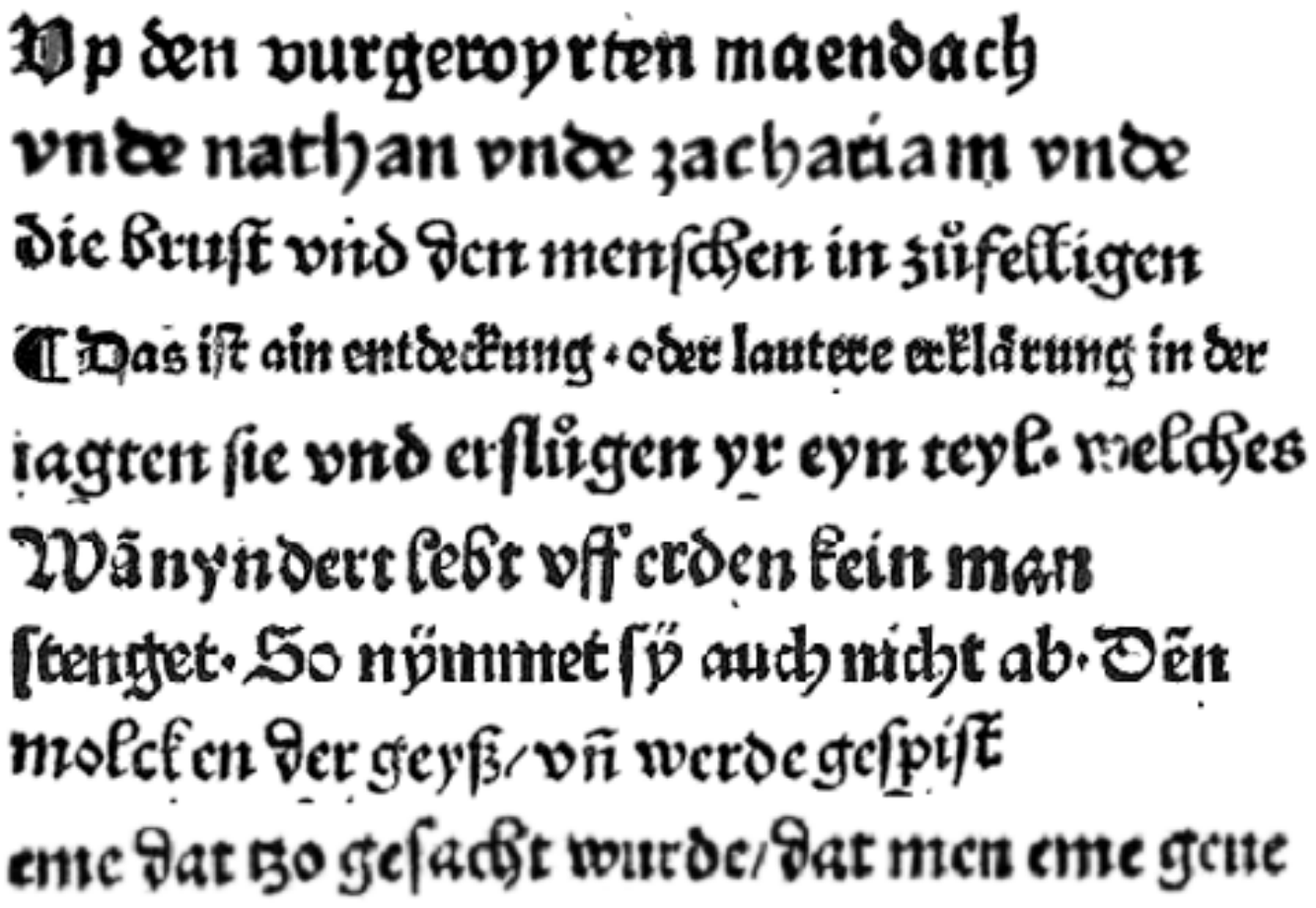

**Figure 2:** Example lines of the Early New High German incunabulum corpus in chronological order (see Table 2).

To get an idea of how well mixed models work for incunabula we trained nine models on eight books each and applied this model to the one book left out of the training set. The resulting CERs are given in the last column of Table 2. The previous finding of Springmann and Lüdeling (2017) that mixed models generalize better than individual models is corroborated: The worst recognition rate is 91.90% with an average rate of 95.40% on unseen books.

We provide a mixed model that was trained on the combined training set of all books and evaluated against a previously unseen test set taken from the same books. The resulting character recognition rate is above 97% for each book in this corpus (a higher value than the previous average because for this model each book contributed to the training set).

### 2.2 The *Kallimachos* corpus

The *Kallimachos corpus* consists of the 1488 printing of *Der Heiligen Leben* and eight books from the *Narragonien digital* subproject.[12] dealing with the second most popular book in its time after the bible, the *Narrenschiff (ship of fools)* by Sebastian Brant. There are four Latin printings (*Stultifera nauis*) translated by Locher and Badius, respectively, two Early New High German printings, one Early Low German work (*Der narrenscip*), and one Latin/English document (Barclay) of which we just provide the Latin part. Whereas the German documents use a broken script, some Latin works are printed with Antiqua types similar to our modern types (Fig. 3). We do not provide a mixed model of these rather diverse types but leave it to the reader to construct his own models for his specific interests. The transcription of Badius is less accurate than that of the other books because it has not yet been checked to the same level of detail.

### 2.3 An *Early Modern Latin* corpus

In Springmann, Fink, and Schulz (2016) we introduced a Latin data set of manual transcriptions from books that were either of interest to us or to scholars who requested an OCR text for a complete book for which we had to train an individual recognition model. The Early Modern Latin corpus is essentially the same, but leaves out the 1497 *Stultifera Nauis* (belonging to the *Kallimachos* corpus) and adds the 1543 *Psalterium* of Folengo (see Table 4). The printings are mostly in Antiqua types (except the *Speculum Naturale* of Beauvais, Fig. 4). The two provided models are those of the above mentioned publication.

[12] http://kallimachos.de/kallimachos/index.php/Narragonien Because annotated transcriptions of the *Narrenschiff* works have not yet been published the single lines of these works have been randomly permutated and do not provide a coherent text in their enumerated order [04.03.2018].

**Table 3:** The Kallimachos corpus

| Year | GW | (Short) Title | # Lines |
|---|---|---|---|
| 1488 | M11407 | Der Heiligen Leben (Winterteil) | 4178 |
| 1495 | 5049 | Das neu narrenschiff | 2114 |
| 1497 | 5051 | Das nuw schiff von narragonia | 1197 |
| 1497 | 5056 | Stultifera nauis | 1424 |
| 1497 | 5061 | Stultifera Nauis | 1092 |
| 1499 | 5064 | Stultifera nauis | 721 |
| 1500 | 5066 | Der narrenscip | 2500 |
| 1505 | | Nauis stultifera (Badius) | 4713 |
| 1509 | | The Shyp of Folys (Barclay) | 2990 |
| | | **Sum:** | **20,929** |

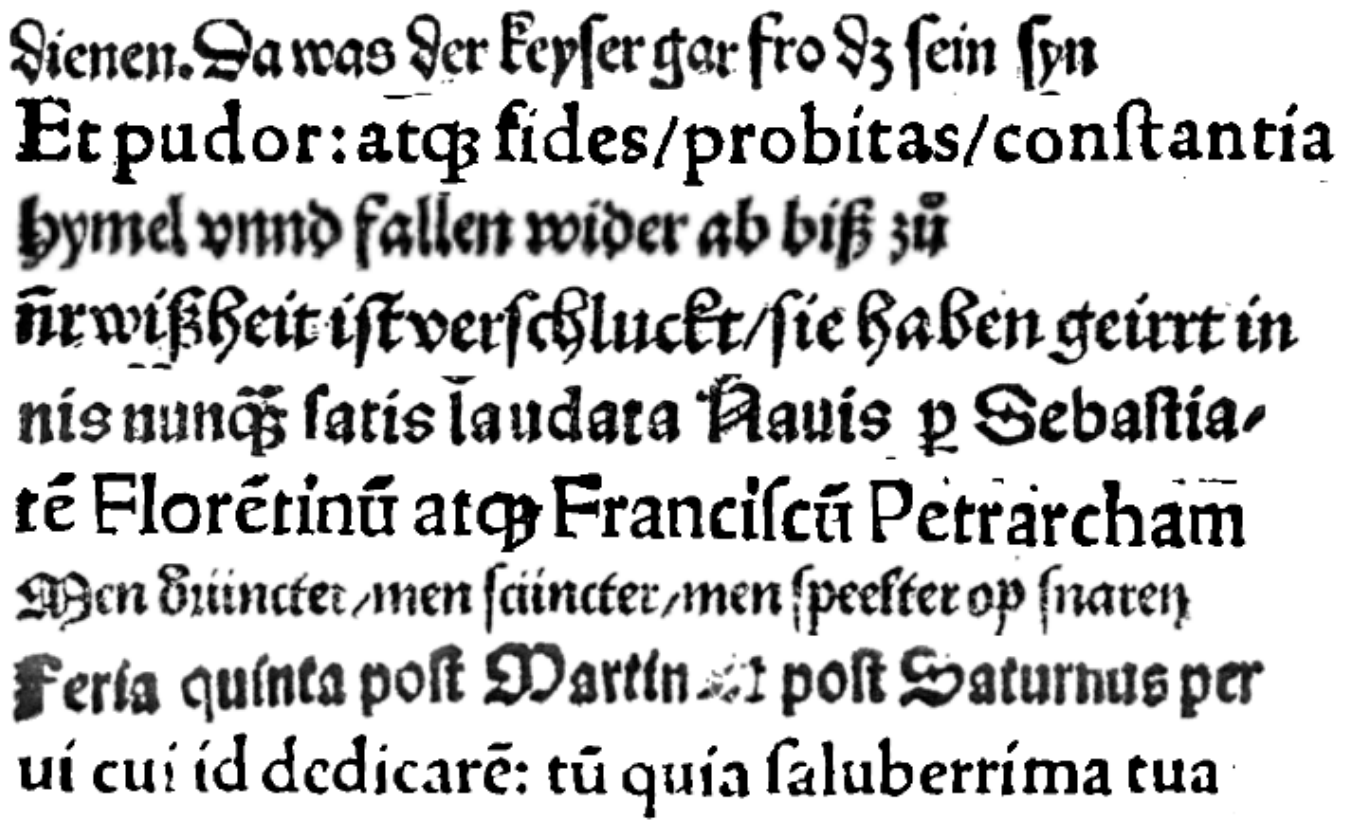
dienen. Da was der keyser gar fro dz sein syn
Et pudor: atq; fides/probitas/constantia
hymel vnnd fallen wider ab biß zů
ůrwißheit ist verschluckt/sie haben geirrt in
nis nunq; satis laudata Nauis p Sebastia-
tē Florētinū atq; Franciscū Petrarcham
Men drincter/men saincter/men speelter op snaren
Feria quinta post Martin post Saturnus per
ui cui id dedicarē: tū quia saluberrima tua

**Figure 3:** Example lines of the Kallimachos corpus in chronological order (see Table 3). Both Antiqua fonts (Latin) and broken fonts (German) are present.

**Table 4:** The Early Modern Latin corpus

| Year | (Short) Title | Author | # Lines |
|---|---|---|---|
| 1471 | Orthographia | Tortellius | 417 |
| 1476 | Speculum Naturale | Beauvais | 2012 |
| 1483 | Decades | Biondo | 915 |
| 1522 | De Septem Secundadeis | Trithemius | 201 |
| 1543 | De Bello Alexandrino | Caesar | 830 |
| 1543 | Psalterium | Folengo | 314 |
| 1553 | Carmina | Pigna | 297 |
| 1557 | Methodus | Clenardus | 350 |
| 1564 | Thucydides | Valla | 1948 |
| 1591 | Progymnasmata vol. I | Pontanus | 710 |
| 1668 | Leviathan | Hobbes | 1078 |
| 1686 | Lexicon Atriale | Comenius | 1216 |
| | | **Sum:** | **10,288** |

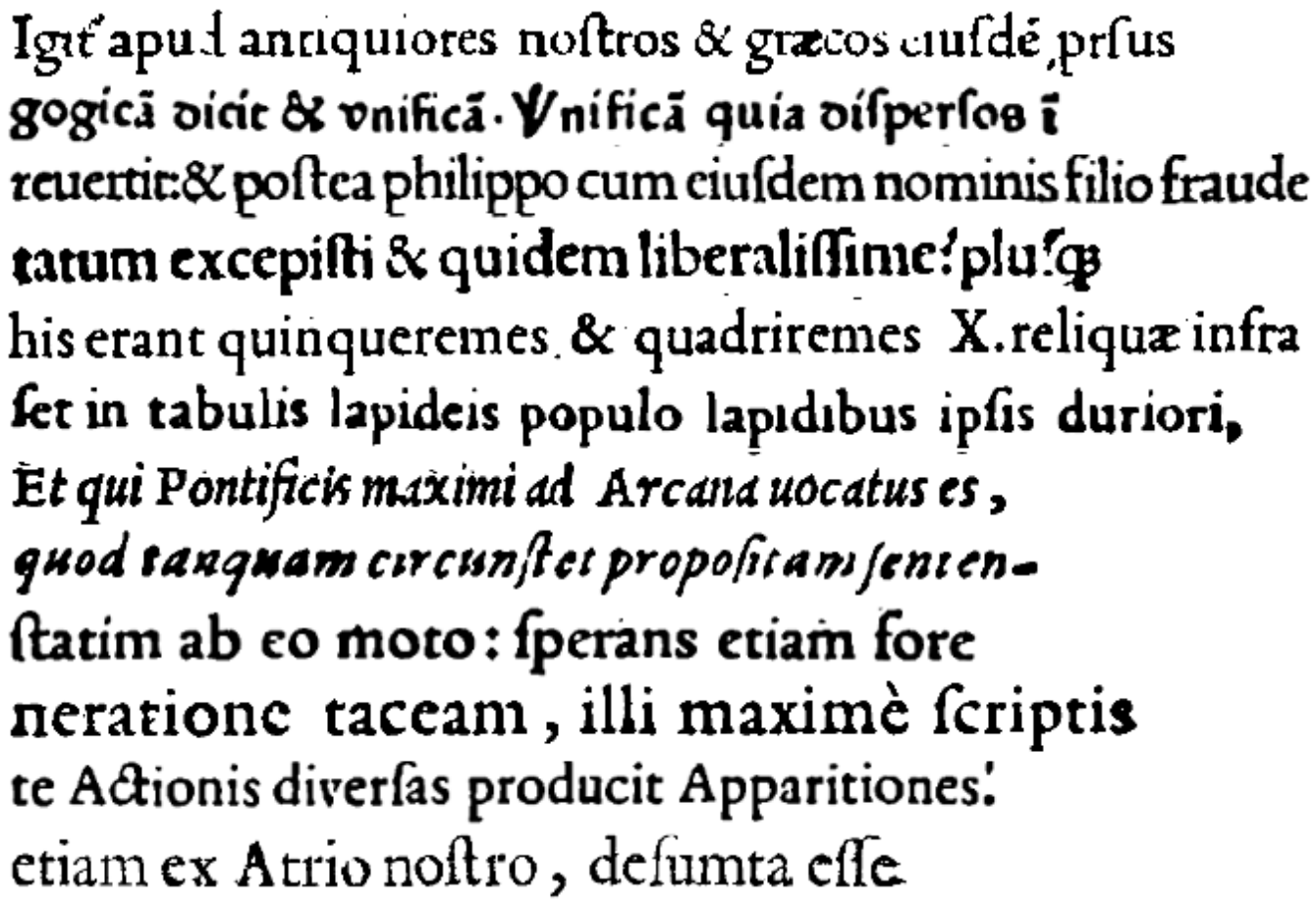
Igit apud antiquiores nostros & graecos eiusdē prsus
gogicā dicit & vnificā. Vnificā quia dispersos ī
reuertit:& postea philippo cum eiusdem nominis filio fraude
tatum excepisti & quidem liberalissime: plusq
his erant quinqueremes. & quadriremes X. reliquae infra
set in tabulis lapideis populo lapidibus ipsis duriori,
Et qui Pontificis maximi ad Arcana uocatus es,
quod tanquam circunstet propositam senten-
statim ab eo moto: sperans etiam fore
neratione taceam, illi maximè scriptis
te Actionis diversas producit Apparitiones:
etiam ex Atrio nostro, desumta esse

**Figure 4:** Example lines of the Early Modern Latin corpus in chronological order (see Table 4).

**Table 5:** The RIDGES Fraktur corpus.

| Year | (Short) Title | Author | # Lines |
|---|---|---|---|
| 1487 | Garten der Gesunthait | Cuba | 747 |
| 1532 | Artzney Buchlein der Kreutter | Tallat | 504 |
| 1532 | Contrafayt Kreüterbuch | Brunfels | 366 |
| 1543 | New Kreüterbuch | Fuchs | 483 |
| 1557 | Wie sich meniglich | Bodenstein | 995 |
| 1588 | Paradeißgärtlein | Rosbach | 795 |
| 1603 | Alchymistische Practic | Libavius | 473 |
| 1609 | Hortulus Sanitatis | Durante | 696 |
| 1609 | Kräutterbuch | Carrichter | 677 |
| 1639 | Pflantz-Gart | Rhagor | 1091 |
| 1652 | Wund-Artzney | Fabricius | 601 |
| 1673 | Thesaurus Sanitatis | Nasser | 733 |
| 1675 | Curioser Botanicus | Anonymous | 567 |
| 1687 | Der Schweitzerische Botanicus | Roll | 520 |
| 1722 | Flora Saturnizans | Henckel | 562 |
| 1735 | Mysterium Sigillorvm | Hiebner | 470 |
| 1764 | Einleitung zu der Kräuterkenntniß | Oeder | 916 |
| 1774 | Unterricht | Eisen | 562 |
| 1828 | Die Eigenschaften aller Heilpflanzen | Anonymous | 658 |
| 1870 | Deutsche Pflanzennamen | Grassmann | 868 |
| | | **Sum:** | **13,248** |

## 2.4 The *RIDGES* Fraktur corpus

The use of broken scripts dates back to the 12th century and was once customary all over Europe. It is therefore of considerable interest to be able to recognize this script in order to OCR the large amount of works printed in a variety of Fraktur. This dataset collects Fraktur material from 20 documents of the *RIDGES* corpus of herbals (Odebrecht et al., 2017) which has been proofread for diplomatic accuracy and matched by us against lines images of the best available scans. OCR experiments on this corpus were reported in Springmann and Lüdeling (2017). The two mixed models used in that publication are provided and give a good base model covering about 400 years of Fraktur printings. Note that the author of the 1543 printing was erroneously attributed to Hieronymous Bock in Springmann and Lüdeling (2017) and has been corrected to Leonhart Fuchs in Table 5.

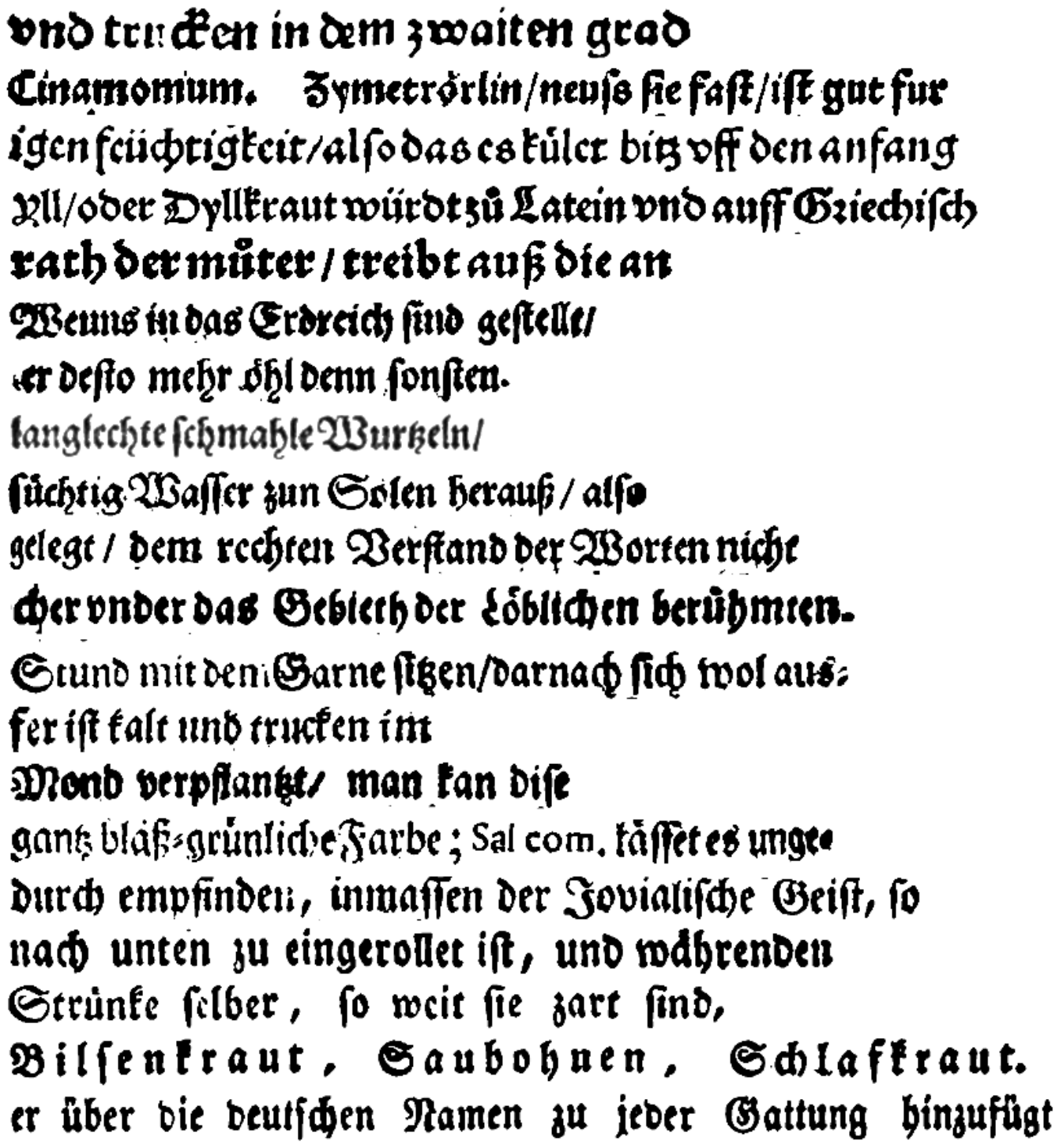
vnd trucken in dem zwaiten grad
Cinamomum. Zymetrörlin/neuſs ſie faſt/iſt gut fur
igen feüchtigkeit/alſo das es kület biß vff den anfang
Yll/oder Dyllkraut würdt zů Latein vnd auff Griechiſch
rath der můter / treibt auß die an
Wenns in das Erdreich ſind geſtellt/
er deſto mehr öhl denn ſonſten.
langlechte ſchmahle Wurtzeln/
ſüchtig Waſſer zun Solen herauß / alſo
gelegt / dem rechten Verſtand der Worten nicht
cher vnder das Gebieth der Löblichen berühmten.
Stund mit dem Garne ſitzen/darnach ſich wol auſ=
fer iſt kalt und trucken im
Mond verpflantzt/ man kan diſe
gantz bläß=grünliche Farbe ; Sal com. läſſet es unge=
durch empfinden, inmaſſen der Jovialiſche Geiſt, ſo
nach unten zu eingerollet iſt, und während
Strünke ſelber, ſo weit ſie zart ſind,
Bilſenkraut, Saubohnen, Schlafkraut.
er über die deutſchen Namen zu jeder Gattung hinzufügt

**Figure 5:** Example lines of the RIDGES Fraktur corpus in chronological order (see Table 5).

## 2.5 The *DTA19* corpus of 19th century German Fraktur

The use of broken scripts in the 19th century and later was mostly restricted to Germany and some neighboring countries. There is a large amount of scans available from 19th century documents (newspapers, long-running journals such as *Die Grenzboten*[13] or *Daheim*, encyclopedias,[14] dictionaries, novels, and reprints of classical works from previous centuries) which are of considerable interest to philologists and historians.

Because of this high interest, some prominent works have been converted into electronic form by manual transcription (keyboarding, double-entry transcription) in low-wage countries.[15] Given the sheer amount of available material, faster and less costly alternatives are sought after and both commercial (ABBYY Finereader with a special Fraktur licence[16]) and open source OCR engines (Tesseract and OCRopus) are capable of recognizing Fraktur printings. What motivated us to look at 19th century Fraktur separately was the question whether we could beat the available general recognition models of the mentioned OCR engines. This is currently an open research topic.

It is tempting to use synthetic training materials, as a variety of Fraktur computer fonts is readily available on the internet. In fact, the Fraktur recognition model of Tesseract is completely based upon synthetic material, the model of OCRopus mostly. However, closer inspection shows that many fonts are either lacking some essential characteristics of real Fraktur types (such as long *ſ*, or *ch* and *tz* ligatures) or have obviously been constructed for calligraphic use and do not reflect the most frequently used historical types. For best OCR results we have to rely on transcriptions of real data, at least as an addition to any synthetic data set one might construct.

In the following we describe a collection of transcriptions from *Deutsches Textarchiv* for which line segmentations from ABBYY Finereader are available. The corresponding scans of these transcriptions are held by *Staatsbibliothek zu Berlin*.[17] We produced line images by cutting page scans into lines using the line coordinates contained in the ABBYY XML output. In this way a corpus of 63 books, some belonging to multi-volume works, could be assembled fully automatically. From these we selected just one volume of each multi-volume edition to provide a balanced multi-font corpus and did some quality checks on correct segmentations by hand.

The resulting *DTA19* corpus of 39 works is detailed in Table 6. To our knowledge there does not exist a similar extensive collection of ground truth for German 19th century Fraktur. We also provide a model trained on this corpus.

Because most Fraktur fonts do not differentiate between the alphabetic characters *I* and *J* and use the same glyph for both, we harmonized the transcription of *DTA* that employs different symbols to just use *J*. Otherwise, a model trained on the original transcription would randomly output either *I* or *J* for the same glyph. As a side effect, however, Roman numerals with the *I* glyph in the image will now be recognized with the *J* letter in the OCR output. This is a systematic error resulting

[13] http://brema.suub.uni-bremen.de/grenzboten

[14] https://www.zedler-lexikon.de/

[15] E.g. Krünitz' Ökonomische Enzyklopädie: http://www.kruenitz1.uni-trier.de/

[16] https://abbyy.technology/en:features:ocr:old_font_recognition

[17] http://staatsbibliothek-berlin.de/

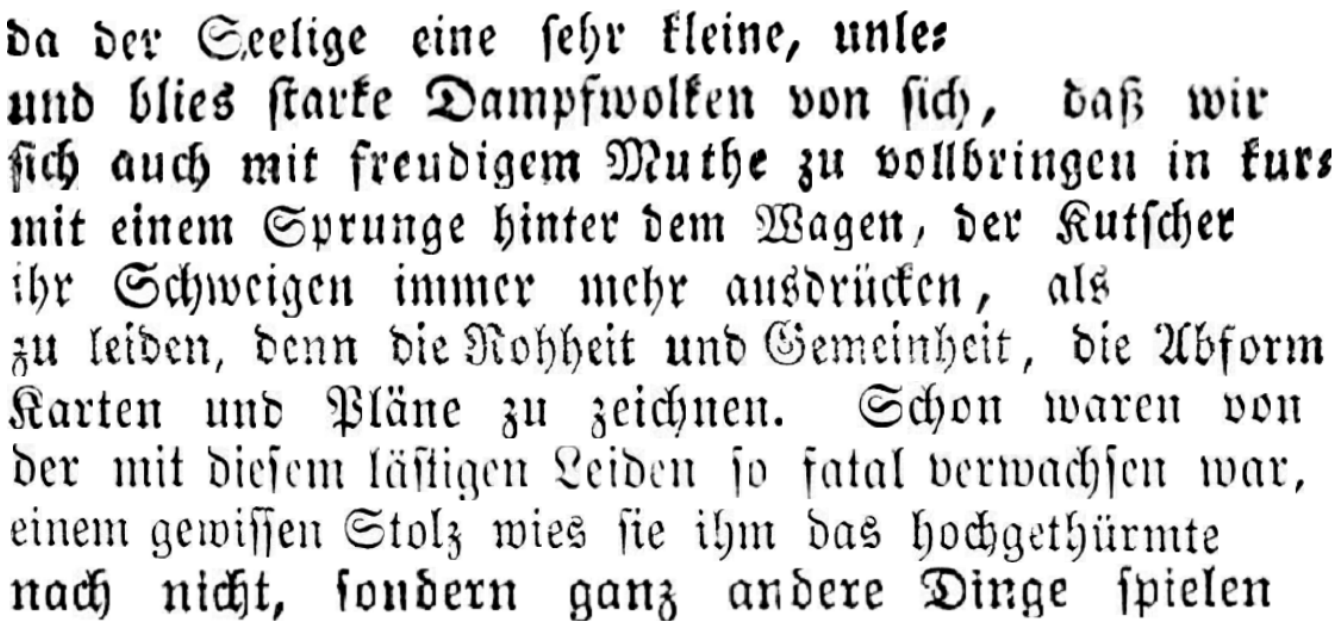
da der Seelige eine ſehr kleine, unle-
und blies ſtarke Dampfwolken von ſich, daß wir
ſich auch mit freudigem Muthe zu vollbringen in kur-
mit einem Sprunge hinter dem Wagen, der Kutſcher
ihr Schweigen immer mehr ausdrücken, als
zu leiden, denn die Rohheit und Gemeinheit, die Abform
Karten und Pläne zu zeichnen. Schon waren von
der mit dieſem läſtigen Leiden ſo fatal verwachſen war,
einem gewiſſen Stolz wies ſie ihm das hochgethürmte
nach nicht, ſondern ganz andere Dinge ſpielen from ground truth that is incorrect for these cases. A better model would result from training on handcorrected ground truth where only Roman numerals have the *I* letter.

**Figure 6:** A selection of lines of the DTA19 corpus. From top to bottom: 1815, 1817, 1819, 1826, 1835, 1853, 1861, 1879, 1891, 1897.

# 3 Other historical ground truth corpora

In the following we mention other historical ground truth corpora which are not part of GT4HistOCR. Only the *Archiscribe* corpus of 19th century German Fraktur is directly usable for OCR model training whereas the others would need various amounts of effort to be aligned as line image/transcription pairs. We also give estimates on the amount of material (number of line pairs) potentially available.

## 3.1 The *Archiscribe* corpus

A prime obstacle for generating ground truth for OCR training purposes consists in the segmentation of textual elements on a printed page into text lines. To circumvent this problem, we made use of several open APIs of the Internet Archive[18] to directly retrieve line images from historical books that can be used as image sources for creating ground truth.

The Internet Archive hosts a collection of over 15 million texts, whose scans are sourced from Google Books as well as a number of volunteers and cooperating institutions.[19] For every scanned book, an automated process creates OCR with ABBYY FineReader. While the actual OCR output of this engine for text with Fraktur typefaces is of very low quality, the resulting line segmentation is usually fairly accurate.

To create ground truth from the Internet Archive corpus, a simple crowd sourcing web application, Archiscribe,[20] is provided. First-time users of the application have to read through a simplified ver-sion of the transcription guidelines of the Deutsches Textarchiv.[21] They are then offered the option to pick a certain year between 1800 and 1900 and set a number of lines they want to transcribe.

[18] `http://archive.org`
[19] `https://archive.org/scanning`
[20] `https://archiscribe.jbaiter.de`; source code: `https://github.com/jbaiter/archiscribe` (MIT license)

**Table 6:** The DTA19 Fraktur corpus.

| Year | (Short) Title | Author | # Lines |
|---|---|---|---|
| 1797 | Herzensergießungen | Wackenroder | 5150 |
| 1802 | Ofterdingen | Novalis | 6198 |
| 1804 | Flegeljahre vol. 1 | Paul | 5332 |
| 1815 | Elixiere vol. 1 | Hoffmann | 8008 |
| 1816 | Buchhandel | Perthes | 861 |
| 1817 | Nachtstücke vol. 1 | Hoffmann | 6578 |
| 1819 | Revolution | Görres | 6178 |
| 1821 | Waldhornist | Müller | 2343 |
| 1826 | Taugenichts | Eichendorff | 7662 |
| 1827 | Liebe | Clauren | 6724 |
| 1827 | Reisebilder vol. 2 | Heine | 5980 |
| 1827 | Lieder | Heine | 5873 |
| 1828 | Gedichte | Platen | 5103 |
| 1828 | Literatur vol. 1 | Menzel | 8124 |
| 1832 | Gedichte | Lenau | 4446 |
| 1832 | Paris vol. 1 | Börne | 5329 |
| 1834 | Feldzüge | Wienbarg | 7805 |
| 1835 | Wally | Gutzkow | 5728 |
| 1852 | Ruhe vol. 1 | Alexis | 9314 |
| 1852 | Gedichte | Storm | 2038 |
| 1853 | Ästhetik | Rosenkranz | 14062 |
| 1854 | Heinrich vol. 1 | Keller | 9343 |
| 1854 | Christus | Candidus | 2095 |
| 1861 | Problematische Naturen vol. 2 | Spielhagen | 6445 |
| 1863 | Menschengeschlecht | Schleiden | 1788 |
| 1871 | Bühnenleben | Bauer | 12008 |
| 1877 | Novellen | Saar | 6354 |
| 1879 | Auch Einer vol. 2 | Vischer | 10492 |
| 1880 | Hochbau | Raschdorff | 661 |
| 1880 | Heidi | Spyri | 6210 |
| 1882 | Sinngedicht | Keller | 11209 |
| 1882 | Gedichte | Meyer | 6262 |
| 1886 | Katz | Eschstruth | 6601 |
| 1887 | Künstlerische Tätigkeit | Fiedler | 4983 |
| 1888 | Irrungen | Fontane | 7079 |
| 1891 | Bittersüß | Frapan | 7008 |
| 1897 | Gewerkschaftsbewegung | Poersch | 1476 |
| 1898 | Fenitschka | Andreas-Salomé | 4753 |
| 1898 | Erinnerungen vol. 2 | Bismarck | 10339 |
| | | **Sum:** | **243,942** |

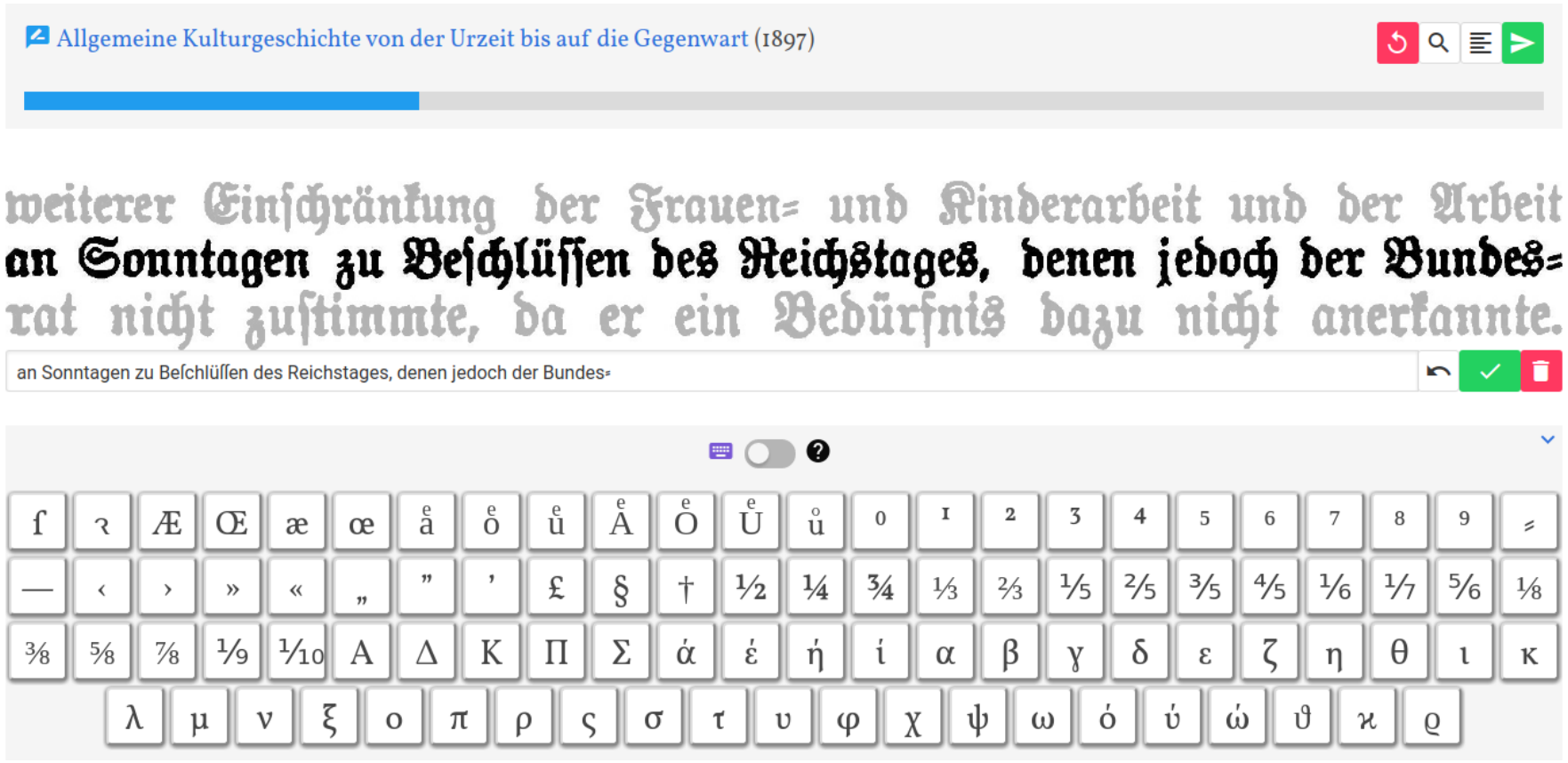

**Figure 7:** Transcribing a line with Archiscribe

In order to retrieve these lines from a suitable book, Archiscribe uses the publicly available search API of the Internet Archive[22] to retrieve a list of 19th century German language texts and randomly picks a volume that has not yet been transcribed. To determine whether a given text is actually set in Fraktur, a heuristic is used: The OCR text is downloaded and searched for the token *ift*, a common misinterpretation by OCR engines trained on Antiqua fonts of the actual word *iſt* (German *ist* = English *is*), which has a high frequency in any German text (of course, real books also contain quotations and other material in Antiqua, as is seen in the second line of Figure 8). If this heuristic results in a false positive (there are some books printed in Antiqua employing a long s), one can just start over. Once a suitable book is found, the desired number of lines[23] are picked at random from the book.

To serve the images to the user, Archiscribe uses the publicly available IIIF Image API endpoint[24] of the Internet Archive. As the API allows the cropping of regions out of a given page image hosted by the archive.org server, the application can directly use it for rendering the line images in the user's browser, and no image processing on the Archiscribe server is neccessary.

Once a suitable volume has been picked and the lines to be transcribed have been determined, the user is presented with a minimal transcription interface consisting of the line to be transcribed, a text box to enter the transcription and an on-screen keyboard with a number of commonly occurring special characters not available on modern keyboards. To offer more context in difficult cases, the user may opt to display the lines above and below the line to be transcribed (Fig.7).

When all lines have been transcribed, they are submitted to the Archiscribe server, where they are stored alongside with their corresponding line images in a Git repository that is published to the corpus repository on GitHub on every change.[25]

[21] http://www.deutschestextarchiv.de/doku/basisformat/transkription.html
[22] https://archive.org/advancedsearch.php
[23] user-defined, by default 50
[24] https://iiif.archivelab.org/iiif/documentation

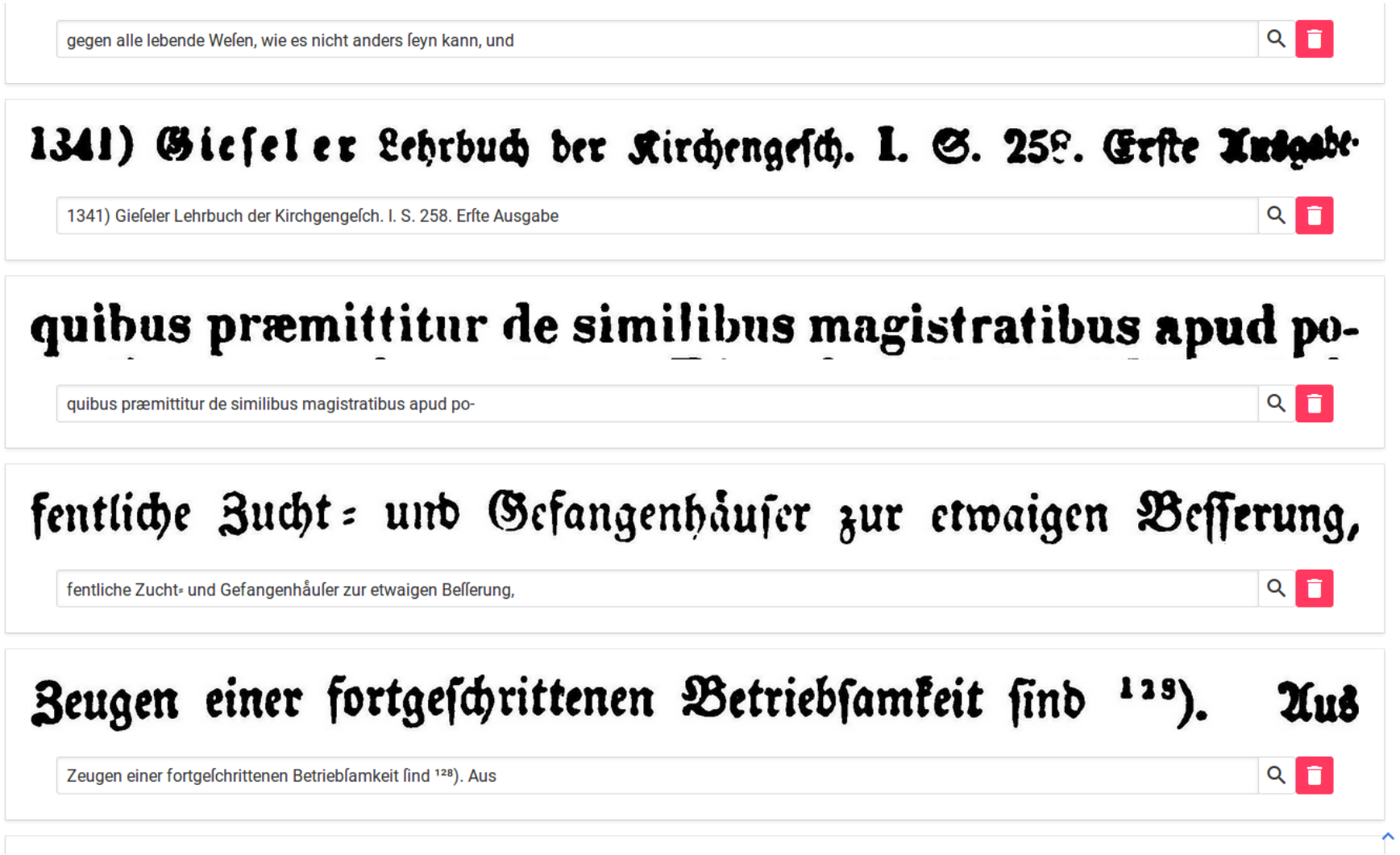

**Figure 8:** Reviewing an existing transcription with Archiscribe. Often books printed in Fraktur also contain lines in Antiqua, mostly quotations in Latin (second line from top). If they are transcribed as well, the model will be able to recognize mixed Fraktur-Antiqua texts.

To ease maintenance of the ground truth corpus a simple review interface is available (Fig.8) where existing transcriptions can be filtered and edited. Due to the use of a Git repository as the storage backend, it is also very easy to keep track of changes in the dataset or to revert some changes in case of vandalism.[26]

Currently the application is restricted to 19th century German language books from the Internet Archive, but it is planned to add support for the transcription of books sourced from any repository that offers a IIIF API, the number of which is steadily increasing.

The *Archiscribe* corpus of ground truth generated by crowdsourcing with the Archiscribe tool currently consists of 4145 lines from 109 works published across 72 years[27] evenly distributed across the whole 19th century. All of the data is available under a CC-BY 4.0 license.

### 3.2 The *OCR-D* ground truth corpus

The OCR-D project funded by Deutsche Forschungsgemeinschaft (DFG) created ground truth of Latin and German printings published between 1500 and 1835 in Germany. This corpus currently consists of one to four pages each of 94 works.[28] Data are provided in both TIFF format (page

[25] https://github.com/jbaiter/archiscribe-corpus
[26] Although the application does not require authentication or registration of any kind, this has not been an issue so far.
[27] [last accessed 31th August 2018]
[28] http://www.ocr-d.de/sites/all/GTDaten/IndexGT.html [last accessed 26 August 2018]

images) and an XML representation in both ALTO and PAGE XML containing the segmentation of the pages in text zones as well as their transcription. In order to produce OCR training data from these files, the text zones of the TIFF images need to be identified by their coordinates contained in the XML files, then these subimages have to be segmented into text lines and matched with the corresponding transcription, also contained in the XML files. We estimate that this dataset currently contains 300 pages and a total of approximately 10,000 lines.

### 3.3 The full *DTA* corpus

There is also the complete DTA corpus of currently 4,422 volumes in German with transcriptions on page level covering the period 1500 to 1900. To produce OCR ground truth fully automatically one needs to segment page images and heuristically match the existing line transcriptions against segmented text line images. Work along this direction is already under way. The amount of available lines is approximately 30 million.[29]

### 3.4 Ground truth from the *IMPACT* project

The EU-funded *IMPACT* project (2008-2012) collected historical ground truth in the form of semantic regions of page images (such as text, image, footnote, marginal notes, page number etc.) for the task of automatic page segmentation (*document analysis*) as well as transcriptions for the text regions of ca. 45,000 pages.[30] Transcribed ground truth is available for several European languages.[31] There may be as many as 1 million lines available, but unfortunately the ground truth comes under a variety of licenses depending on the contributing institution and can currently only be downloaded page by page.

## 4 Notes on transcription guidelines for OCR

To produce training data for OCR where a machine will decide what label to attach to a printed glyph, the golden rule is: *The same glyph must have the same transcription, even if the glyph has different context dependent meanings*. Otherwise, the machine will get confused and randomly output one of the different characters or character sequences it has learnt to associate with the glyph. Consequently the single Fraktur glyph for the letters *I* and *J* can only have one character representation, not two, and ambiguous and context dependent abbreviations must not be resolved. E.g., a vowel with tilde above in Early Modern Latin could either mean (vowel+m) or (vowel+n). A further example is provided by the r-hook above letter *d* in Table 7. Also, ignoring line endings of printed lines and merging hyphenated words will destroy the correspondence between printed line image and transcription needed for model training.

[29] http://www.deutschestextarchiv.de/doku/ueberblick
[30] https://www.primaresearch.org/www/media/datasets/ImpactRepositoryPoster.pdf
[31] See https://www.digitisation.eu/

This makes most of the existing transcriptions of historical documents which resolve abbreviations, merge hyphenated words at line endings, correct printing errors, and modernize historical spellings unusable for OCR purposes. What is needed instead is a *diplomatic transcription*, i.e. a transcription of printed glyphs to characters with no or minimal editorial intervention.[32]

But even if we transcribe diplomatically, there is still room for a decision on the level of detail we want to transcribe, e.g. if we want to record the usage of long s (ſ) or rounded r (*r rotunda*). The collection of explicit recordings of such decisions are called transcription guidelines. They are indispensable to ensure a consistent text, both over time and between different people transcribing parts of same document. They are also necessary if you want to pool data from different corpora which have been transcribed by different guidelines. You have to inspect the guidelines in order to regularize different data sets to a common norm.

Explicit transcription guidelines exist for the Reference Corpus ENHG,[33] texts from DTA,[34] and RIDGES.[35] All other corpora had to be made internally consistent with our Perl script. The correctness of the data will determine the predictive power of any machine model trained on it. We define the correctness of a transcription as its adherence to predefined, internally consistent transcription guidelines and not as the level of detail which it records. We emphasize this point because we have been set back by inconsistent data produced by researchers, students and the public alike.

Note that a linguistically motivated transcription (such as in the *Reference Corpus Early New High German* or the *Deutsches Textarchiv*) might very well choose to transcribe similar looking glyphs by differently looking characters for a specific use case such as search. In order to use these transcriptions for OCR model training one needs to normalize to just one alternative (*J*, in our case). Examples of differences between a linguistic transcription and a transcription for OCR training are shown in Table 7 for the Reference Corpus ENHG.

## 5 Conclusion

Historical OCR has been advanced to a state where even very early printings from the 15th century can be recognized by individually trained models with a character recognition rate of 98% and above. To be practical on a large scale, however, pretrained models are needed that result in recognition rates >95% without any prior training requirement. As long as we lack an automatic method to revive historical fonts to build large synthetic corpora the construction of pretrained models rests on the availability of historical ground truth. The *GT4HistOCR* dataset is put forward to allow experimentation and research under a permissive CC-BY 4.0 license and is a first step for the construction of widely applicable pretrained models for Latin and German Fraktur. We hope that other researchers will follow our example and make their ground truth available under an open source license in directly usable form for OCR training.

[32] `http://www.stoa.org/epidoc/gl/latest/trans-diplomatic.html`
[33] Not yet publically available
[34] See Footnote 22
[35] `https://www.linguistik.hu-berlin.de/de/institut/professuren/korpuslinguistik/forschung/ridges-projekt/documentation/download-files/pubs/ridgesv8_2018-04_06.pdf`, pp. 248 ff.

**Table 7:** Extract from the transcription guidelines of the Reference Corpus ENHG. The transcript column shows examples of the linguistically motivated transcription, the UTF-8 column represents our interpretation for OCR purposes.

| Example | Transcript | UTF-8 | Description |
|---|---|---|---|
| 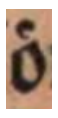 | o\e | oͤ | vowel modifier |
| 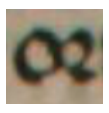 | o_r | or | ligature |
| 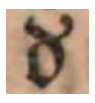 | d’ | ꝺ | d with abbreviation of <er>, <r>, <ir>, <re>, <ri> |
| 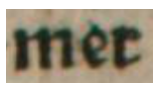 | me<t> | met | letters that are difficult to read |
| 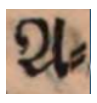 | A= | A⸗ | word-internal line break |

## Acknowledgments

We are grateful to our colleagues Phillip Beckenbauer for aligning the ground truth of the *Early New High German Corpus* to the printed text lines of the respective books and the training of a mixed model, to the *Narragonien digital* project (Joachim Hamm/Brigitte Burrichter) providing corrected transcriptions as ground truth, especially by Christine Grundig and Thomas Baier and their collaborators, and to the many students and collaborators of the *RIDGES Corpus* at Humboldt University Berlin. Uwe Springmann produced most of the Early Modern Latin Corpus, Johannes Baiter is the author of the Archiscribe tool. The Kallimachos project has been funded by BMBF under grant no. 01UG1415A and 01UG1715A.